\documentclass[letterpaper]{article} 
\usepackage{aaai23-r2hcai}  
\usepackage{times}  
\usepackage{helvet}  
\usepackage{courier}  
\usepackage[hyphens]{url}  
\usepackage{graphicx} 
\usepackage{natbib}  
\usepackage{caption} 
\usepackage{algorithm}
\usepackage{algorithmic}

\usepackage{newfloat}
\usepackage{listings}
\DeclareCaptionStyle{ruled}{labelfont=normalfont,labelsep=colon,strut=off} 
\floatstyle{ruled}
\newfloat{listing}{tb}{lst}{}
\floatname{listing}{Listing}
\usepackage{fancyhdr}
\fancypagestyle{firstpagehf}
{
   \fancyhf{}
   \fancyhead[HC]{The AAAI 2023 Workshop on Representation Learning for Responsible Human-Centric AI (R$^2$HCAI)}
   \fancyfoot[C]{\thepage}
}

\title{Dive into the Resolution Augmentations and Metrics in Low Resolution Face Recognition: A Plain yet Effective New Baseline}
\author {
    Xu Ling,\textsuperscript{\rm 1}
    Yichen Lu,\textsuperscript{\rm 1}
    Wenqi Xu,\textsuperscript{\rm 1}
    Weihong Deng,\textsuperscript{\rm 1}
    Yingjie Zhang,\textsuperscript{\rm 1,2}
    Xingchen Cui,\textsuperscript{\rm 1,2}
    Hongzhi Shi,\textsuperscript{\rm 1,2}
    Dongchao Wen \thanks{Corresponding author}\textsuperscript{\rm ,1,2}
}
\affiliations {
    \textsuperscript{\rm 1} Inspur Electronic Information Industry Co., Ltd\\
    \textsuperscript{\rm 2} Shandong Massive Information Technology Research Institute, Jinan, China\\
    \{lingxu421, wenqixu99\}@gmail.com, yichen.iu@outlook.com, dengweihong2020@foxmail.com, \{zhangyj-s, cuixingchen, shihzh, wendongchao\}@inspur.com
}

\usepackage{bibentry}

\usepackage{amsmath}
\usepackage{booktabs}
\usepackage{multirow}
\usepackage{subfig}

\DeclareUnicodeCharacter{2061}{}
\begin{document}
\thispagestyle{firstpagehf}
\maketitle

\begin{abstract}
    Although deep learning has significantly improved Face Recognition~(FR), dramatic performance deterioration may occur when processing Low Resolution~(LR) faces. 
    To alleviate this, approaches based on unified feature space are proposed with the sacrifice under High Resolution~(HR) circumstances. 
    To deal with the huge domain gap between HR and LR domains and achieve the best on both domains, we first took a closer look at the impacts of several resolution augmentations and then analyzed the difficulty of LR samples from the perspective of the model gradient produced by different resolution samples.
    Besides, we also find that the introduction of some resolutions could help the learning of lower resolutions. 
    Based on these, we divide the LR samples into three difficulties according to the resolution, and propose a more effective Multi-Resolution Augmentation.
    Then, due to the rapidly increasing domain gap as the resolution decreases, we carefully design a novel and effective metric loss based on a $LogExp$ distance function that provides decent gradients to prevent oscillation near the convergence point or tolerance to small distance errors; it could also dynamically adjust the penalty for errors in different dimensions, allowing for more optimization of dimensions with large errors.
    Combining these two insights, our model could learn more general knowledge in a wide resolution range of images and balanced results can be achieved by our extremely simple framework.
    Moreover, the augmentations and metrics are the cornerstones of LRFR, so our method could be considered a new baseline for the LRFR task.
    Experiments on the LRFR datasets: SCface, XQLFW and large-scale LRFR dataset: TinyFace demonstrate the effectiveness of our methods, while the degradation on HRFR datasets is significantly reduced.
\end{abstract}

\section{Introduction}
The high accuracy of High Resolution Face Recognition~(HRFR) tasks mainly attributed to large-scale face data~\cite{webface,cao2018vggface2} collected in uncontrolled scenarios and various softmax-based loss functions~\cite{liu2016large,liu2017sphereface,wang2018additive,deng2019arcface}. 
However, due to the huge domain gap between LR and HR faces and very limited LR faces data in the wild, the performance of the Low Resolution Face Recognition~(LRFR) tasks is still poor.
Furthermore, LRFR tasks are again widely seen in real scenarios, e.g. LR faces captured from surveillance cameras at long distances, which also push the problem to be improved~\cite{fang2020generate}.

\begin{figure}[t]
  \centering
  \includegraphics[width=1.0\linewidth]{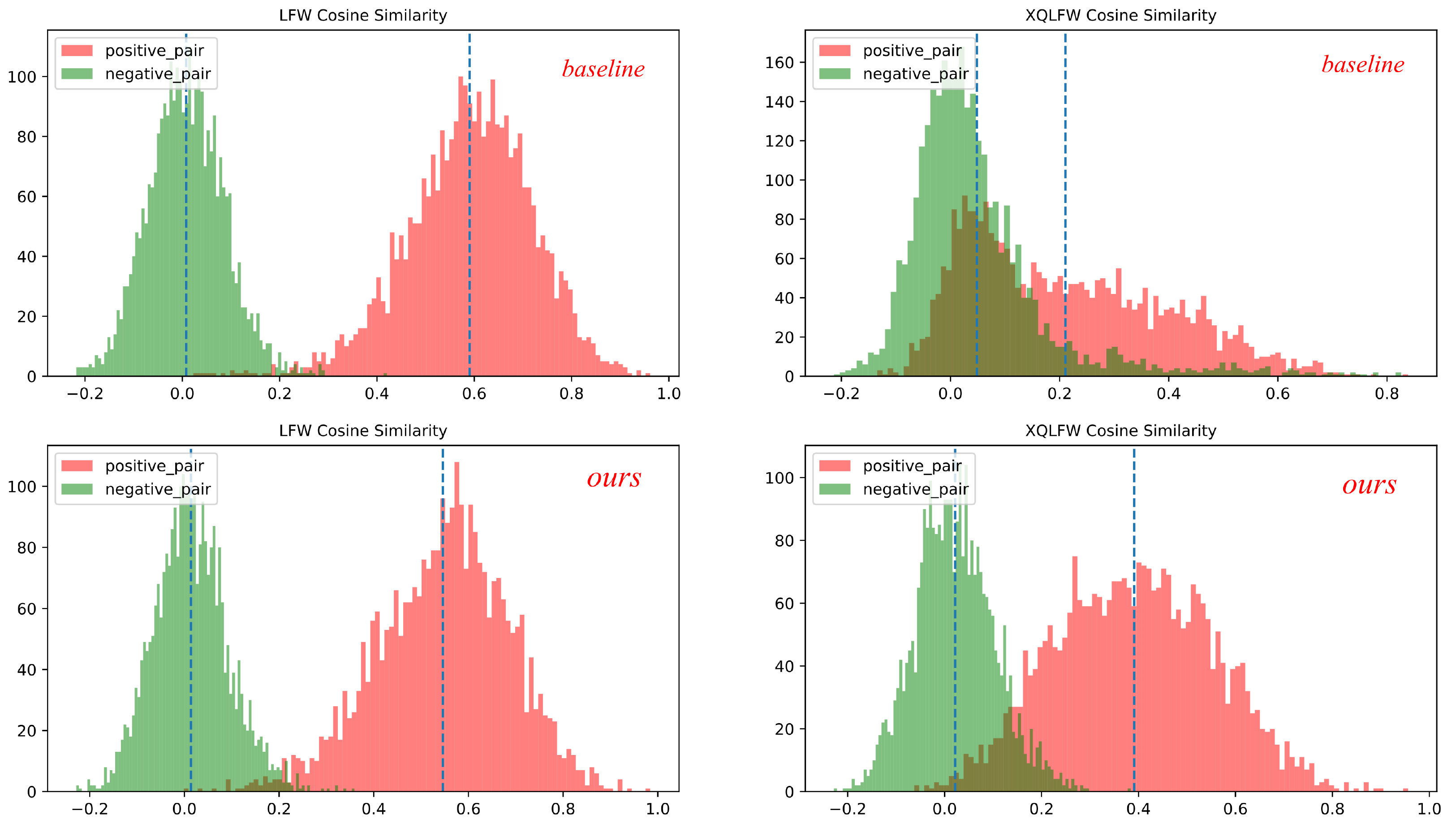}

  \caption{The negative and positive cosine similarity distribution of LFW~\cite{LFWTech} and XQLFW~\cite{knoche2021crossquality} datasets, which represent the HR and LR cases correspondingly. 
  In comparison to ours, we used the HR model as a baseline.
  For the LR case, our model can extract identity information from the faces, thus reducing the overlap of cosine similarity of positive and negative pairs. At the same time, the HR case has only been slightly harmed.}
  \label{fig:one}
\end{figure}

As shown in Figure~\ref{fig:one}, if the HR model is directly applied to recognize LR faces, a large portion of the positive pairs will have close to orthogonal cosine similarity, while a small portion of the negative pairs will have very comparable features. 
This indicates that the information loss caused by the resolution and massive domain gap between LR and HR mislead the model to fail to extract the correct identity from the faces in the LR domain. 
Therefore, it becomes crucial to fuse the two widely disparate domains so that the model can be resolution-independent, and inadequate and inappropriate fusion can lead to undesirable side effects of reduced HR performance~\cite{hong2019unsupervised,shi2020towards}.

In general, data augmentation is the most common tool to resolve data insufficiency, and Multi-Resolution augmentations are commonly used in LRFR.
However, previous work tend to simply select some resolutions for augmentation in equal proportions~(ST-M~\cite{knoche2021image}:$7px$,$14px$,$28px$,$56px$, Bridge KD~\cite{ge2020efficient}:$16px$,$32px$,$64px$,$96px$, etc). They lack detailed study and analysis, and thus the settings may not be optimal.
Therefore, we conduct some experiments to verify the importance of different resolutions.
We first tested models with different resolution augmentation on synthetic LFW at different resolutions to analyze the effect of different resolution augmentation. 
More specifically, we found that the HR model could be robust to relatively high resolution inputs~(from 28~$px$ to 112~$px$) and that even a 28~$px$ augmentation did not result in a significant performance gain.
We also found that low-resolution augmentations can bring performance gains~(from 7~$px$ to 14~$px$), however, due to the large gaps between domains, introducing them directly is often not optimal.
At the same time, augmentation using multiple different resolutions gives better results than using only one resolution. 
Further, to maximum the potential of Multi-Resolution Augmentation~(MAug), we use model back propagation gradient to analyze the learning difficulty of different resolution samples and then we give the relationship between the difficulty of the synthesized samples and the resolutions. 
In light of these, we classify the synthetic low-resolution faces into \emph{extremely hard}, \emph{hard} and \emph{semi-hard} samples based on their resolution and propose a new multi-resolution augmentation. Significant performance improvements could be obtained.

Furthermore, from the model gradient we also found that there is learning migration between resolutions, i.e. the introduction of some resolutions in training could reduce the learning difficulty of samples with lower resolutions.
Therefore, with proper Multi-Resolution Augmentation, a decent metric loss could leverage the gradient information to guide the model to gradually learn lower resolutions samples during training.
We first analyzed the $L_1$ and $L_2$ losses from the gradient perspective. It may be found that due to the constant gradient value, $L_1$ loss leads to the oscillation problem as the two embeddings are unable to converge further when they are relatively close. 
On the other hand, $L_2$ loss fails to provide enough gradient when the distance is relatively close, which could lead to tolerance of tiny distance and insufficient convergence.
These convergence difficulties can lead to inadequacies in fusing the two domains.
So, we proposed a new metric loss $L_{LogExp}$, which could avoid neglecting the tiny distance and mitigate the oscillation problem, resulting in a more effective feature extractor.

In summary, the main contributions of this paper are as follows:
\begin{itemize}
  \item We dive into the resolution augmentations to verify the importance of different resolutions and find the learning migration between resolutions.

  \item We introduce the concept of hard samples into resolution augmentation and propose a new multi-resolution augmentation method that only augments at low resolutions.

  \item We review the popular distance function from the perspective of gradient. Then, we propose a novel distance loss to better fuse the widely separated HR and LR domains, which is simple to implement and can be integrated with a variety of other approaches.
  
  \item Our proposed method can be considered a new baseline for the LRFR task, and it can be conveniently integrated into other methods. 
  Meanwhile, we also conduct a large number of experiments and the state-of-the-art results in LRFR are achieved while the HRFR performance degradations are reduced.  
\end{itemize}

\section{Related Work}

\subsection{Common Face Recognition}

Softmax-based loss~\cite{deng2019arcface,wang2018additive,liu2017sphereface,liu2016large}, as well as metric learning loss~\cite{schroff2015facenet,sun2015deep}, have substantially improved general face recognition(or HR face recognition)~\cite{wang2018survey}. 
Meanwhile, large-scale face datasets such as MillionCelebs~\cite{Zhang_2020_CVPR} and Webface42M~\cite{zhu2021webface260m} serve as the foundations for face recognition to improve. 
However, since the detailed information in LR faces is less than in HR faces, resulting in a massive domain gap between HR and LR, the HR model's performance under LR circumstances would suffer significantly.

\subsection{Low Resolution Face Recognition}

\textbf{SR-based Method.}
The Super Resolution(SR)-based approach expects to fill the lost information to generate synthetic HR images~\cite{zangeneh2020low,chen2018fsrnet,ataer2019verification,mudunuri2018genlr}.
To restrict the features of the synthetic images and HR images, \citeauthor{zhang2018super} proposed Super-Identity loss. Fang et al.~\cite{fang2020generate} suggested using MR-GAN  to generate realistic LR images for training, avoiding the issue of unrealistic downsampling to some extent.

\textbf{Feature-based Method.}
However, the quality of the synthesized HR images is difficult to be proven valid and may even carry the bias of the SR model, while the SR method increases the computational effort. Thus, on the other hand, feature-based works try to learn projection into a unified domain and minimize the distances between LR and HR embeddings.
\citeauthor{lu2018deep} proposed a deep coupled ResNet (DCR) model that employs branch networks to transform the image embeddings for each domain to learn robust features. 
Ge and Massoli et al.~\cite{ge2020look,ge2020efficient,ge2018low,zhao2019low,massoli2020cross} expect to achieve better LR performance by using knowledge distillation to migrate helpful knowledge for LR from the HR model to the student network. 
\citeauthor{li2021deep} proposed using Rival Penalized Competitive Learning to drive the most similar class center away to learn more discriminative features, and high performance was attained with such a simple concept. 
\citeauthor{singh2021derivenet} recently proposed replacing the softmax loss with Derived-Margin Softmax Loss, which may alter the margin dynamically during training.

\begin{figure*}[t]
  \centering
  \includegraphics[width=0.75\linewidth]{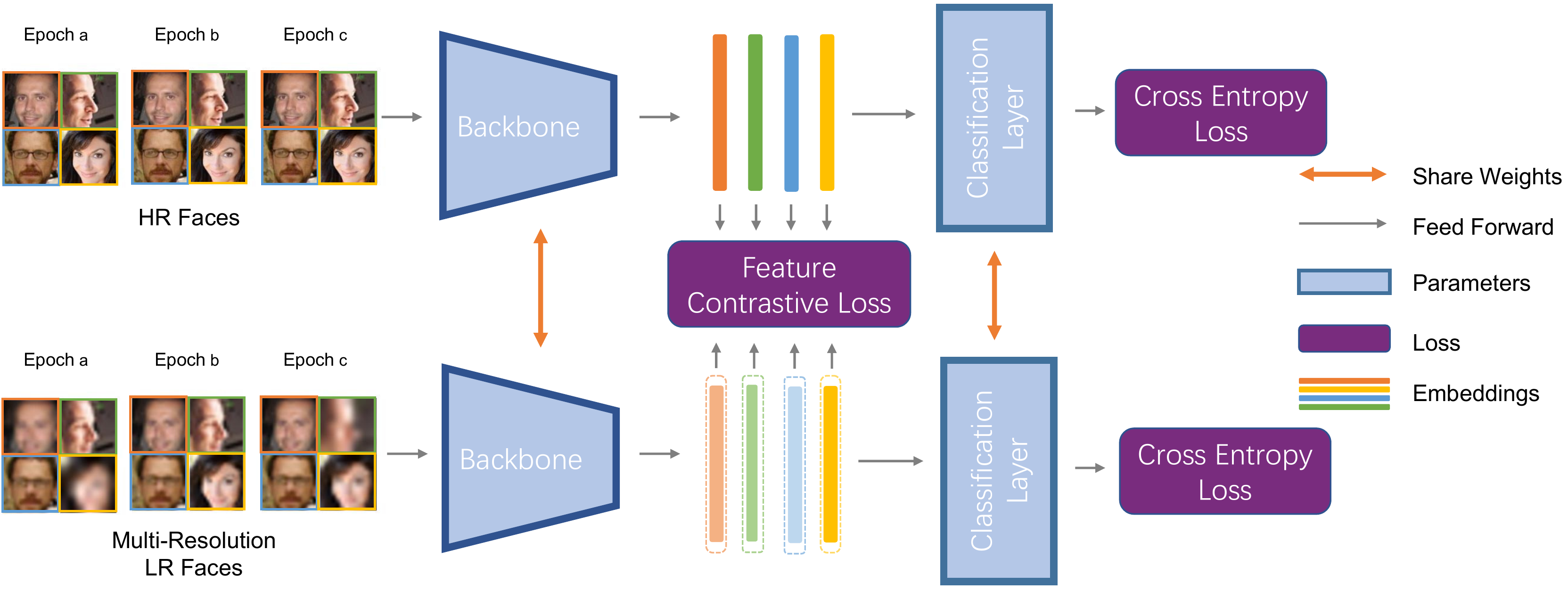}

  \caption{Overview of our proposed method. The current input HR faces and the multi-resolution LR faces are input into the backbone to get the corresponding embeddings. The Feature Contrastive Loss is used to constrain the distance. 
  CosFace is used to ensure that embeddings are classified to the correct identity.}
  \label{fig:method}
\end{figure*}

\textbf{Previous Augmentations.}
Although LR faces can be obtained by GAN-based methods~\cite{li2021deep} or Gaussian blurring methods~\cite{shi2020towards}, interpolation-based methods are the most popular (bicubic is most commonly-used method). They first down-sampling the input to a certain resolution and then up-sampling it to fit the input dimension. 
Previous work uses a particular resolution (\Citeauthor{zhao2019low}:$32px$), or multi-resolution such as DCR:$40px$,$6px$, ST-M \cite{knoche2021image}:$7px$,$14px$,$28px$,$56px$, Bridge KD \cite{ge2020efficient}:$16px$,$32px$,$64px$,$96px$, etc.
However, Multi-Resolution Augmentation has not been carefully studied in the previous works. 
In this paper, we study in detail the relationship among resolution, performance and model gradient, and propose a novel and effective multi-resolution augmentation based on bicubic interpolation.

\subsection{Contrastive Learning and Siamese Network}

Our research takes a deep dive into the resolution augmentations and metric functions in LRFR and demonstrates their enormous potential. Therefore, we did not introduce a complex network structure, but resorted to the most naive structure in the LRFR method based on the unified feature space~\cite{mudunuri2018genlr}: the Siamese network structure. 
At the same time, the structure adopted in this paper is similar to contrastive learning. 
In self-supervised learning, contrastive-based algorithms have also achieved competitive results or even surpassed the performance of supervised learning in several tasks~\cite{oord2018representation,he2020momentum,chen2020simple}.
However, we does not introduce negative samples, which is the main difference from contrastive learning.
In summary, we adopt the most intuitive, plain structure to demonstrate the effectiveness and superiority of our proposed technique.

\section{Methodology}

Guided by the Siamese network, our subsequent research and exploration are based on this most concise structure.
The overview of our proposed method is illustrated in Figure~\ref{fig:method}.
In this paper, we propose \textbf{1. a new Multi-Resolution Augmentation} and \textbf{2. a new Feature Contrast Loss}.

\subsection{Multi-Resolution Augmentation}
\label{multiresolu}

The resolution augmentations use interpolation method~(e.g.BICUBIC) to downsample and then upsample HR images to reduce the image resolution but keep the input dimension.
In this section, we mainly focus on three experiments to answer two questions: 1. In multi-resolution augmentation, which resolutions should be augmented, and 2. Why they are effective.

\begin{figure*}[h]
  \centering
  \subfloat[]{\includegraphics[width=0.33\textwidth]{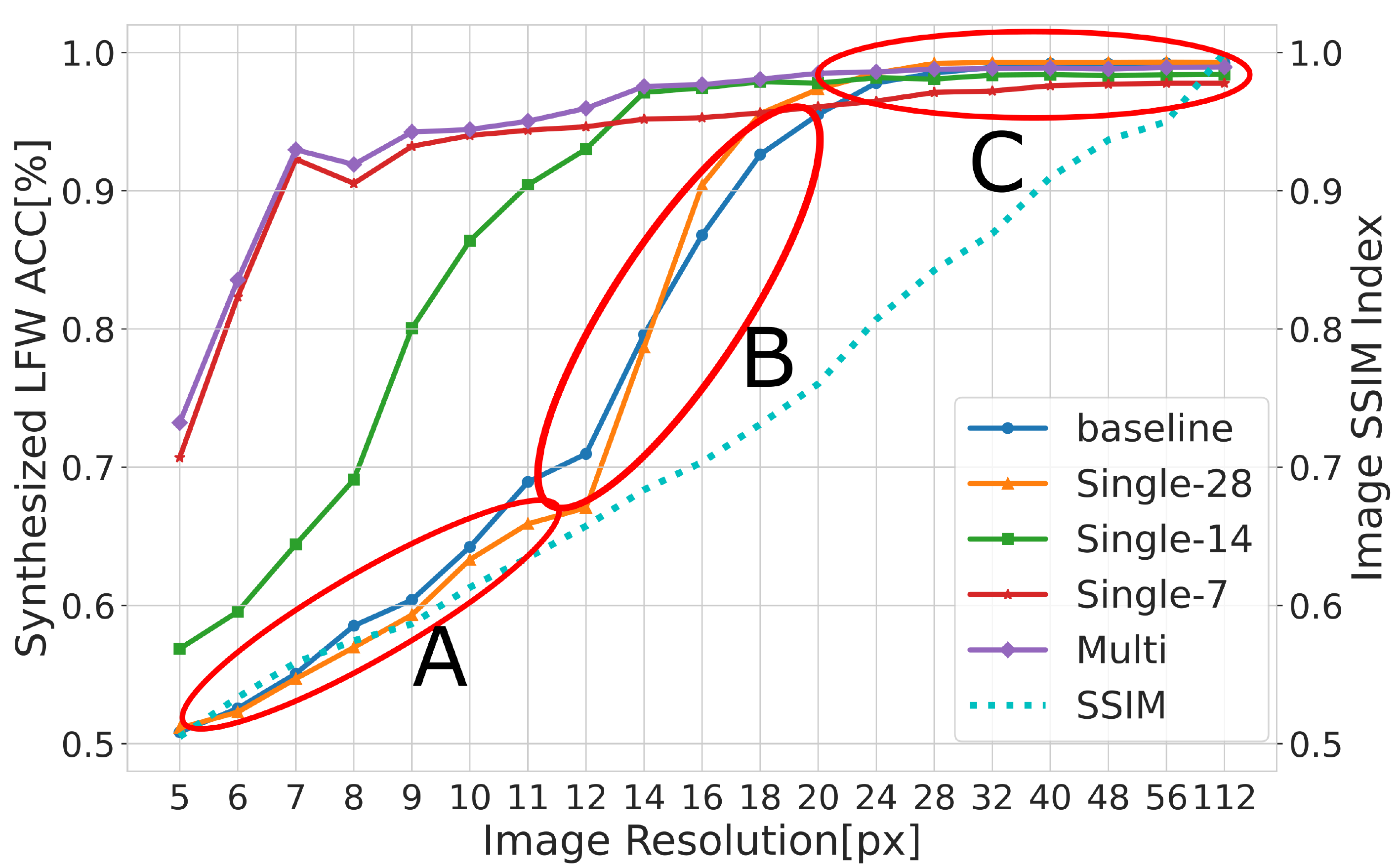}\label{fig:three}}
  \subfloat[]{\includegraphics[width=0.33\textwidth]{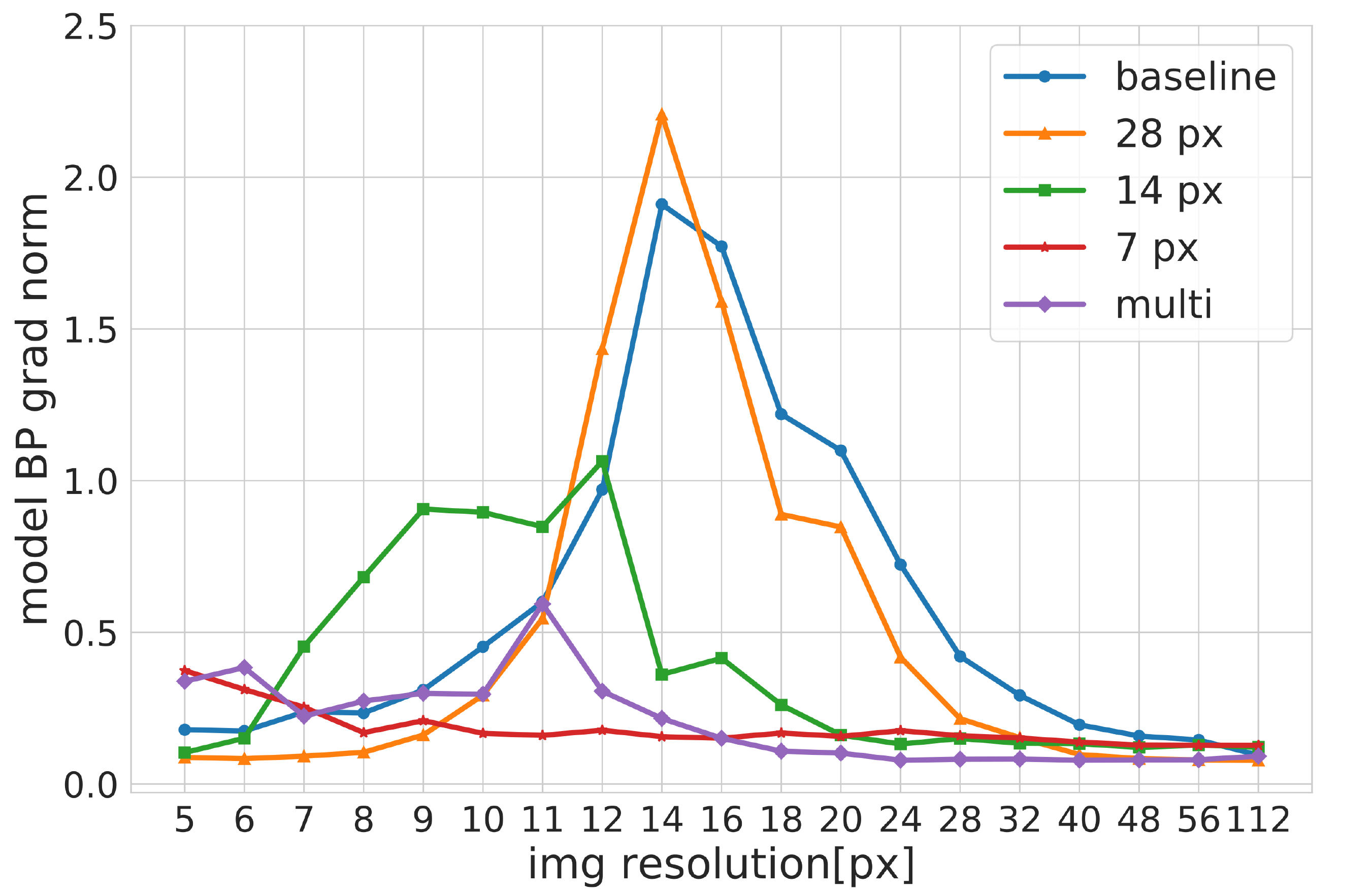}\label{fig:grad_lfw_px}}
  \subfloat[]{\includegraphics[width=0.33\textwidth]{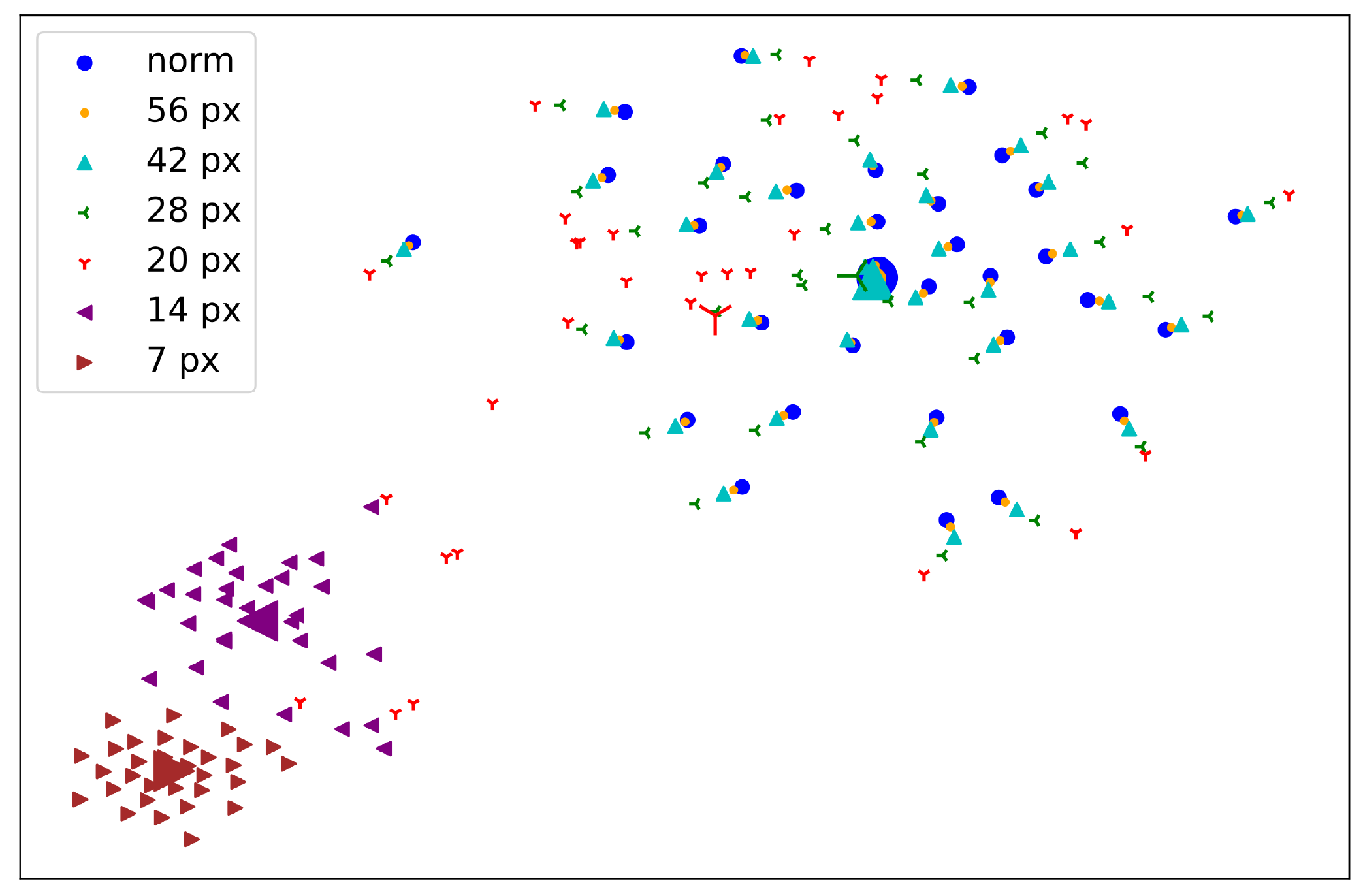}\label{fig:tsne_base}}

  \caption{(a):Synthesized LFW accuracy for different image resolutions with our $L_{LogExp}$.
  \emph{Single-x} denotes that only $x~px$ and HR faces are utilized to train the model. \emph{Multi} indicates that multiple resolution images are used during training where we employed the 7~$px$,14~$px$,20~$px$ and HR faces, while the training set size remains the same as \emph{Single-x}. 
  The general HR model is used as \emph{baseline}. The SSIM Index for different resolutions are also shown.
  (b):The gradient norm sum of the parameters of the model in (a) after back-propagation for different resolution inputs.
  (c):The t-SNE result of the HR model~(Baseline in (a)) for the features of the same faces at different resolutions, the bigger point represents the center of each resolution.}
\end{figure*}

First, we analyze the effect of resolution augmentations when a feature contrast loss is used between the HR and the synthesized LR faces. As shown in Figure~\ref{fig:three}, we test models with various resolution augmentations on synthetic LFW at different resolutions and give the SSIM Index~\cite{wang2004image} of different resolution images. 
Secondly, we compute the gradient sums for different models in Figure~\ref{fig:three} for different inputs, as shown in Figure~\ref{fig:grad_lfw_px}.
We also provide the t-SNE~\cite{van2008visualizing} results of the HR model for the features of the same faces at 7 representative different resolutions, as shown in Figure~\ref{fig:tsne_base}.
Accordingly, we classify the difficulty of different resolution samples and propose a new Multi-Resolution Augmentation. 

As the 112~$px$ could be the most common input size in FR. 7~$px$ and 14~$px$, as the 16x and 8x scaling of 112~$px$, is widely considered in Multi-Resolution Augmentations \cite{knoche2021image}.
However, as for the 28~$px$, which was also selected as one of the augmentation resolutions by previous work, we argue that its improvement in performance is marginal.
It could be found from Figure~\ref{fig:three} that the HR model is robust to relatively high resolution images~(from 28~$px$ to 112~$px$) and its performance only starts to drop dramatically when the resolution is lower than 28~$px$. 
In addition, the use of 28~$px$ resolution augmentation, as shown in Figure~\ref{fig:three}, does not greatly help the model to improve its performance.
The same conclusion can be drawn from Figure~\ref{fig:tsne_base}, where 
the features are significantly shifted starting from 20~$px$ and up to 14~$px$ and 7~$px$. 
These indicate that relatively high resolution images are not completely unfamiliar to the HR model, which we believe mainly thank to the presence of some LR images in the original large-scale training data.
Therefore, the relatively high resolution augmentations~(bigger than 28~$px$) commonly used in previous methods do not need to be introduced.

However, for the introduction of augmentations with a resolution of 7~$px$ or 14~$px$ alone, although performance gains at low resolutions are obtained as shown in Figure~\ref{fig:three}, performance deterioration occurs at high resolutions compared to the HR model. 
Similarly, the feature points in Figure~\ref{fig:tsne_base} with resolution above 28~$px$ could always find their corresponding HR faces, while when the resolution is less than 20~$px$, this correspondence of the same image is not obvious. And even the 14~$px$ and 7~$px$ features form their own clusters. 
These indicates that the loss of information increases rapidly as the resolution decreases; meanwhile, the domain gap between low resolutions may be greater than those between low and high resolutions, so the direct introduction of low resolution will make the HR performance undergo a serious degradation. 

One option for improving the performance of resolution augmentation is to use multiple resolutions simultaneously, and we here give a possible explanation of why multi-resolution augmentation is more effective.
When inputting different samples, the model will derive different gradients to optimize the loss, and the norm of the gradient can determine the magnitude of the amount of change the samples make to the model parameters. The norm of the gradient is also often used in hard mining~\cite{li2019gradient}. 
As shown in Figure~\ref{fig:grad_lfw_px}, the gradient of the models shows a trend of increasing and then decreasing with decreasing resolution.
Also, we found some migration between resolutions: the introduction of some resolutions shifted the peak of the gradients to a smaller resolution. This indicates that the introduction of some resolutions will make it less difficult to learn at lower resolutions (7$px$-14$px$ is much easier for the Single-14 model than Baseline, as in Figure~\ref{fig:three}\&\ref{fig:grad_lfw_px}).
That's why Multi-Resolution could obtain a better result. Different but proper resolution could be the bridge to fuse the HR and LR domains well.

On the other hand, due to the huge domain gap between 14~$px$ and 112~$px$, directly introducing 7~$px$ and 14~$px$ as Multi-Resolution Aug would cause the model to suffer too much on HR~(as shown in Table~\ref{tab:ablation2}); at the same time, due to the aforementioned migration between resolutions, choosing a resolution between 14~$px$ and 112~$px$ for augmentation could better help the model's learning.
It is obvious from Figure~\ref{fig:grad_lfw_px} that 20~$px$ is the point where the feature center starts to shift significantly. So based on the above three points, we also introduce 20~$px$ into our multi-resolution augmentation as well.

Finally, as for the learning difficulty of different resolutions, one could figure out from Figure~\ref{fig:three}\&\ref{fig:grad_lfw_px} that that the model gradient and performance are negatively correlated in the 14$px$-112$px$ region.
However, when the resolution is less than 14~$px$, the gradient of the model is small and the performance is also low in Figure~\ref{fig:three}, indicating that this part of the samples are extremely hard samples which cannot produce enough gradients, making optimization challenging.
Therefore based on our experiments and findings above, we argue that samples below 12~$px$ should be considered as \emph{extremely hard} samples (A in Figure~\ref{fig:three}), and combined with Figure~\ref{fig:tsne_base}, one can consider 20~$px$ to 32~$px$ as \emph{semi-hard} samples (C in Figure~\ref{fig:three}), and 12~$px$ to 20~$px$ as \emph{hard} samples (B in Figure~\ref{fig:three}).
Thus, we propose to combine three different difficulties in training, using three representative resolutions of 7~$px$, 14~$px$, and 20~$px$ for augmentation.

\subsection{Feature Contrast Loss}

Due to the huge domain gap between LR and HR domains, the feature contrast loss need to be carefully designed.
However, we find that the commonly-used $L_1$ and $L_2$ losses exhibit similar performance under our framework~(As shown in Table~\ref{tab:ablation}), which indicates that neither is optimal. 
We thus propose a novel Feature Constraint Loss $L_{LogExp}$ which incorporates the advantages of both based on the analysis of their merits.

\subsubsection{Analysis of Existed Distance Functions.}

First, for $x$ and $y$ are the two vectors we want to pull together, $L_1$ loss:
\begin{equation}
  L_1(x,y) = \frac{1}{D}  \sum_{i=1}^D |x_i-y_i|, \qquad \frac {\partial L_1}{\partial{x_i}}    = \frac{1}{D} \cdot (\pm 1),
  \label{eq:L1}
\end{equation}
gives the same penalty to all distance errors in all dimensions~($|\frac {\partial L_1}{\partial{x_i}}|=\frac{1}{D}$).
The errors in different dimensions tend to be different. Therefore, when we have optimized the dimensions with relatively small errors by $L_1$, the dimensions with large errors need further optimization, which leads to the oscillation problem near the convergence point.
On the other hand, although $L_2$ loss:
\begin{equation}
  L_2(x,y)  = \frac{1}{2D}  \sum_{i=1}^D (x_i-y_i)^2, \qquad \frac {\partial L_2}{\partial {x_i}}  = \frac{1}{D} \cdot (x_i-y_i),
  \label{eq:L2}
\end{equation}
treats different dimensions based on their own distance errors, tolerance of small distance errors may occur because of the small gradient of $L_2$~($|\frac {\partial L_2}{\partial {x_i}}|=\frac{1}{D} \cdot |x_i-y_i|$) when the two embeddings are relatively close, which is also disappointed.
The $SmoothL_1$ function, as a combination of $L_1$ and $L_2$ would also encounter the same problem as $L_2$.

\subsubsection{Our $LogExp$ Distance Function.}

\begin{figure}[t]
  \centering
  \includegraphics[width=0.8\linewidth]{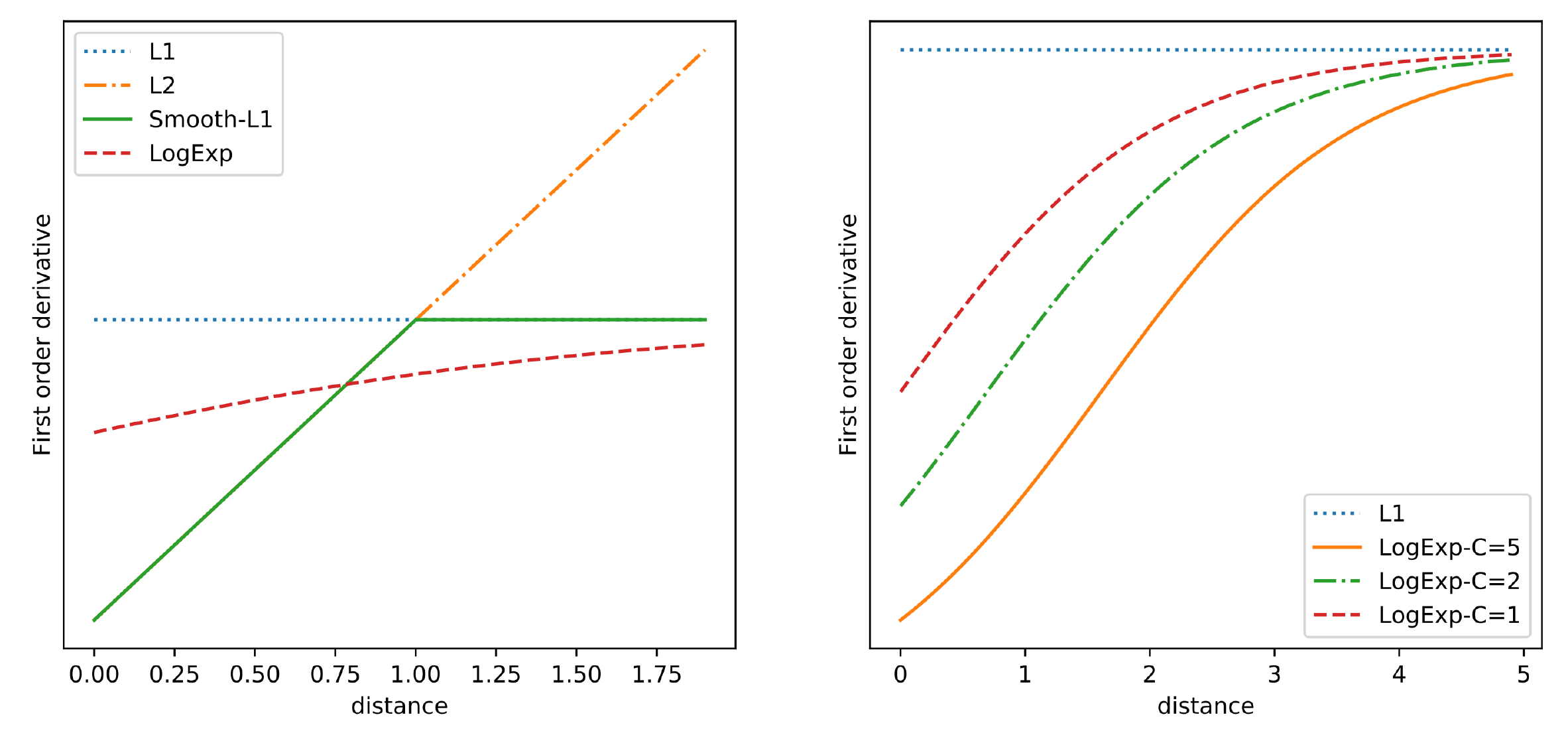}

  \caption{Left:The First-order derivative values of the four distance functions. 
  Right:The effect of different $C$ on the first-order derivatives of the $L_{LogExp}$ functions in Equation~\ref{eq:Llogexp_grad_re}.
  Our $L_{LogExp}$, could not only provide different gradients according to the error on the current dimension~(Left), but also adjust the gradient magnitude according to the ratio of errors between dimensions~(Right).}
  \label{fig:deri}
\end{figure}

Therefore, based on the analysis above, a distance function that not only provides different gradients considering the errors in different dimensions~(Advantage of $L_2$), but also provides gradients with sufficient magnitude when the two embeddings are close to one another~(Advantage of $L_1$), would be the proper one. 
Additionally, due to the direct constraint between embeddings, we also expect its gradient to be smooth.
Here we give a novel Feature Constraint Loss $L_{LogExp}$ with the formula and gradient: 
\begin{equation}
  L_{Log⁡Exp}=\frac{1}{D}\log⁡(1+\sum_{i=1}^D( e^{|x_i-y_i|} -1)),   
  \label{eq:Llogexp}
\end{equation}
\begin{equation}
  \frac{\partial L_{Log⁡Exp}}{\partial{x_i}}=\frac{1}{D} \cdot \frac{(\pm 1) \cdot e^{|x_i-y_i|}}{1+\sum_{i=1}^D( e^{|x_i-y_i|} -1)}.      
  \label{eq:Llogexp_grad}
\end{equation}
The first-order gradient of $L_{LogExp}$ is shown in Figure~\ref{fig:deri}.
Also, just like both $L_1$ and $L_2$ could be generalized as: $L_p  = \frac{1}{p \cdot D}  \sum_{i=1}^D (|x_i-y_i|)^p$,
$L_p$ Loss is $L_1$ loss when $p=1$ and is $L_2$ loss when $p=2$, we could also reference this idea to give $L_{LogExp}$ in general form as:
\begin{equation}
  L_{Log⁡Exp}=\frac{1}{p\cdot D}log⁡(1+\sum_{i=1}^D( e^{(|x_i-y_i|)^p} -1)). 
  \label{eq:Llogexpp}
\end{equation}

\subsubsection{More Advantage of $L_{LogExp}$.} The above gradient formula Equation~\ref{eq:Llogexp_grad}, could be rewritten by considering the terms unrelated to the current dimension as a remainder term $C$, and thus can be rewritten as :

\begin{equation}
  \frac{\partial L_{Log⁡Exp}}{\partial{x_i}}=\frac{1}{D} \cdot \frac{(\pm 1) \cdot e^{|x_i-y_i|}}{C+e^{|x_i-y_i|}}.      
  \label{eq:Llogexp_grad_re}
\end{equation}
As shown in Equation~\ref{eq:Llogexp_grad} and Figure~\ref{fig:deri}, our novel distance function $L_{LogExp}$ gives an intermidiate gradient between the $L_1$ and $L_2$.
Similarly, the remainder term $C$ in Equation~\ref{eq:Llogexp_grad_re} represents the sum of distance in the other dimensions, and the effect of $C$ is shown in Figure~\ref{fig:deri}. 
When $C$ is large, the gradient value of the current dimension becomes smaller, while when $C$ is small, the gradient value of the current dimension becomes larger. 
This can be considered that the $L_{LogExp}$ function could dynamically adjust the gradient magnitude according to the ratio of errors between dimensions.
Thus, it is possible to focus on the dimensions with larger errors during the training process to improve this part first.

\subsection{Total Loss}

Cross-entropy loss with class label can push negative pairs apart, thus we only employ contrast loss to pull positive pairs together.
Therefore, we could simply use the contrast loss to constrain the distance between the corresponding HR and LR faces and make them classify to the right class with the help of Softmax-based loss, e.g. CosFace~\cite{wang2018additive}, to learn discriminative knowledge. 

So, our proposed training process and formulation are as follows: $I_{HR}$ represents the current input HR face from a large-scale dataset, e.g. WebFace~\cite{webface}. 
First, a resolution is randomly selected from the multi-resolution list and then $I_{HR}$ is augmented with BICUBIC to obtain the corresponding $I_{LR}$. 
Thereafter, a convolutional neural network (CNN) backbone~\cite{he2016deep} is used to obtain the embeddings $f$ of $I_{HR}$ and $I_{LR}$, $f_{HR} =F_{CNN}(I_{HR})$ and $f_{LR} =F_{CNN}(I_{LR})$. 
Then CosFace~\cite{wang2018additive} is used as the classification loss $L_{cls}$ and combines the proposed contrastive loss $L_{LogExp}$ to constrain the two embeddings after normalization. 
The final total loss obtained is

\begin{equation}
  \begin{aligned}
    L_{dist} &=L_{Log⁡Exp}(\frac{f_{HR}}{||f_{HR}||_2}, \frac{f_{LR}}{||f_{LR}||_2}), \\
    L_{all}  &= \lambda \cdot L_{dist}+\frac{1}{2}L_{cls}(f_{HR})+\frac{1}{2}L_{cls}(f_{LR}).
  \end{aligned}
  \label{eq:Lall}
\end{equation}

\section{Experiments}

\subsection{Dataset and Training Settings}

\subsubsection{Dataset.}
We chose the large-scale face datasets CASIA-WebFace~\cite{webface} and the cleaned version~\cite{deng2019arcface} of MS1M \cite{ms1m} as training dataset. 
For the test set, SCface~\cite{grgic2011scface} is used as the common cross-resolution recognition dataset, which contains 130 ids and 4,160 face images. 
XQLFW~\cite{knoche2021crossquality} is a recently proposed challenging cross-resolution dataset that solves the problem of non-uniformity of the synthetic LFW datasets used in previous work~\cite{li2021deep}.
TinyFace~\cite{cheng2018low} is also used to test the performance on large-scale low resolution faces, which contains 169,403 LR faces.
We also used LFW~\cite{LFWTech} to measure the performance degradation under HR scenarios.

\subsubsection{Implementation Details.}
In the training process, the resolution and ratio of 7~$px$: 14~$px$: 20~$px$ = 1:1:2 are used for multi-resolution augmentation.
Our resolution augmentation is based on BICUBIC.
Random Horizontal Flip is used for data augmentation. 
All training data were aligned and cropped to 112$\times$112 utilize MTCNN~\cite{MTCNN}. 
SGD momentum was used as the optimizer for training and the momentum was set to 0.9, and weight decay was 5e-4.
When CASIA-WebFace~\cite{webface} was used as training set, ResNet34 \cite{he2016deep} was used as the backbone network with initial learning rate was set to 0.1, a total of 28 epochs were trained and the learning rate was divided by 10 at 16 and 24 epochs. 
When MS1M was used as the training set, ResNet50~\cite{he2016deep} was used as the backbone, the initial learning rate was set to 0.01, a total of 16 epochs are trained and the learning rate is divided by 10 at 8 and 12 epochs. 
The $\lambda$ is set to 1. The $s$ in CosFace loss is set to 48 and $m$ is set to 0.4. We perform our experiments on two GTX1080TIs, the implementation is based on PyTorch~\cite{NEURIPS2019_9015} and the batch size is set to 256.\footnote[1]{ The code is released at: https://github.com/hurricanelx/LRFR }

\begin{table}[t]
  \centering
  \begin{tabular}{l|rrr|r}
    \toprule
    Methods & \multicolumn{1}{l}{$d1$} & \multicolumn{1}{l}{$d2$} & \multicolumn{1}{l|}{$d3$} & \multicolumn{1}{l}{Avg} \\
    \hline
    DCR~\shortcite{lu2018deep}   & 73.3  & 93.5  & 98.0  & 88.27  \\
    TCN~\shortcite{zha2019tcn}   & 74.6  & 94.9  & 98.6  & 89.37  \\
    DATL~\shortcite{ghosh2019learning}  & 76.2  & 96.9  & 98.1  & 90.40  \\
    DAQL~\shortcite{ghosh2019learning}  & 77.3  & 96.6  & 98.1  & 90.66  \\
    RAN~\shortcite{fang2020generate}   & 81.3  & 97.8  & 98.8  & 92.63  \\
    DDL(Res50)~\shortcite{huang2020improving} & 86.8  & 98.3  & 98.3  & 94.47  \\
    NPT-Loss(Res50)~\shortcite{khalid2022npt} & 85.7  & \emph{99.1}  & 99.1  & 94.63 \\
    RPCL-DCR~\shortcite{li2022deep} & 90.4  & 98.0  & 98.0  & 95.47  \\
    \hline
    Ours(w/o FT) & 73.0  & 96.0  & 98.8  & 89.27  \\
    Ours(Res34)  & \emph{91.8} & 99.0 & \emph{99.3} & \emph{96.67} \\
    Ours(Res50)  & \textbf{92.5} & \textbf{99.3} & \textbf{99.5} & \textbf{97.08} \\
    \bottomrule
    \end{tabular}%
    \caption{Rank-1 face identification accuracy~(\%) on the SCface~\cite{grgic2011scface}; 'w/o FT' denotes straight testing with the trained model without fine-tuning.}
    \label{tab:scface}%
\end{table}%

\subsection{Compare With Other Methods}

\subsubsection{Results On SCface.}

Following~\cite{lu2018deep,li2019low,yang2017discriminative} 80 of the 130 ids were used for testing, and the remaining 50 ids that did not intersect were utilized to fine-tune the entire network. 
The results of our method and the comparison with the previous method are shown in Table~\ref{tab:scface}. 
From this, it could figure out that, after fine-tuning, our method is superior to all prior methods, and reaches the state-of-the-art results.
It's also worth noting that, our Res34-based model outperforms DDL~\cite{huang2020improving} and NPT-Loss~\cite{khalid2022npt} despite having ResNet50 as their backbone. 
Furthermore, the performance of our method without fine-tuning is slightly outperformed the DCR~\cite{lu2018deep} technique, i.e., our model could do even better than DCR~\cite{lu2018deep} despite not having seen the SCface data. 
As for the generative based method~\cite{fang2020generate}, our method could also demonstrate better performance.

\subsubsection{Results On TinyFace.}

TinyFace \cite{cheng2018low} is a large-scale low resolution face dataset of 169,403 LR faces that were captured under surveillance conditions. 
Following the same setting with previous work for a fair comparison, we use the training set of TinyFace to fine-tune our trained model before 1:N identification tests. The test results are shown in Table~\ref{tab:tiny}.
Our method provides outstanding results, RPCL enhances CosFace's result by 3\%, while our method improves it by 10\%.
It's also worth noting that despite having much more complex loss functions and twice as many parameters, the URL's~\cite{shi2020towards} result is only 4\% better than our ResNet50-based method.

\begin{table}[t]
  \centering
  \begin{tabular}{l|rrrr}
    \toprule
    Methods & \multicolumn{1}{l}{R-1} & \multicolumn{1}{l}{R-5} &\multicolumn{1}{l}{R-10} & \multicolumn{1}{l}{R-20} \\
    \hline
    DCR~\shortcite{lu2018deep} & 0.29 &  -  & 0.40 & 0.44 \\
    PeiLi's~\shortcite{li2018face} & 0.31 &  -  & 0.43 & 0.46 \\
    Cheng's~\shortcite{cheng2018low} & 0.36  & -   & 0.46 & 0.50 \\
    CosFace(Res34) & 0.29  & -   & 0.39  & 0.42 \\
    RPCL(Res34)~\shortcite{li2022deep} & 0.32  & -   & 0.41  & 0.44 \\
    CosFace(Res100) & 0.47  & 0.52   & -  & - \\
    URL(Res100)~\shortcite{shi2020towards}& \textbf{0.63}  & \textbf{0.68}   & -  & - \\
    \hline
    Ours(Res34)  & 0.39  & 0.46 & 0.50   & 0.54 \\
    Ours(Res50)  & \textbf{\emph{0.59}}  & \textbf{\emph{0.64}} & \textbf{0.68}   & \textbf{0.73} \\
    \bottomrule
  \end{tabular}
  \caption{Face identification Rank-1, Rank-5, Rank-10 and Rank-20 accuracy on TinyFace. R-1 represent for Rank-1.}
  \label{tab:tiny}%
\end{table}%

\subsubsection{Results On XQLFW.}

\begin{table}[t]
  \centering
  \resizebox{\linewidth}{!}{
  \begin{tabular}{l|rlr}
    \hline
    Models & \multicolumn{1}{l}{LFW} & XQLFW & \multicolumn{1}{l}{Avg} \\
    \hline
    CosFace(Res34)~\shortcite{wang2018additive} & 99.42  & 68.95(-30.47) & 84.19  \\
    ArcFace(Res50)~\shortcite{deng2019arcface} & 99.50  & 74.22(-25.28) & 86.86  \\
    MagFace(Res50)~\shortcite{meng2021magface} & 99.63  & 76.95(-22.68) & 88.29  \\
    QMagFace-50~\shortcite{terhorst2021qmagface} & \textbf{99.73} & 80.63(-19.1) & 90.18 \\
    BT-M(Res50)~\shortcite{knoche2021image}  & 99.30  & 83.6(-15.70) & 91.45  \\
    ST-M2(Res50)~\shortcite{knoche2021image} & 95.87  & 90.82(-5.05) & 93.35  \\
    Face Transformer~\shortcite{zhong2021face} & 99.70  & 87.9(-11.80) & 93.80  \\
    ST-M1(Res50)~\shortcite{knoche2021image} & 97.30  & 90.97(-6.33) & 94.14  \\
    \hline
    Ours(Res34) & 99.03  & 92.18(-6.85) & 95.61  \\
    Ours(Res50) & 99.40 & \textbf{94.33}(-5.07) & \textbf{96.87}  \\
    \hline
    \end{tabular}%
    }
    \caption{Face verification accuracy~(\%) for several previous approaches on LFW and XQLFW. Most compared results are directly cited from~\cite{knoche2021crossquality}. 
    }
    \label{tab:xqlfw}%
\end{table}%

The results are presented in Table~\ref{tab:xqlfw}.
From the results, for previous LRFR methods, e.g. ST-M1, ST-M2~\cite{knoche2021image}, although it improves the performance on the XQLFW dataset significantly compared to HR methods~(ArcFace)~(from 75\% to 91\%), it also faces the degradation on LFW~(from 99.5\% to 96\%-97.3\%). 
However, as demonstrated in Table~\ref{tab:xqlfw}, our Res50-based method improves accuracy by 20.11\% on XQLFW compared to ArcFace baseline and just decays 0.10\% on LFW. In contrast, compared to the prior technique ST-M1~\cite{knoche2021image}, which reached the optimum on XQLFW, our method improves 2.31\% on XQLFW and 3.36\% on LFW simultaneously.
That is, our approach could improve the performance of the LRFR task while leaving the LFW performance mostly unaffected, which is both difficult and essential.

\begin{figure}[t]
  \centering
  \includegraphics[width=1.0\linewidth]{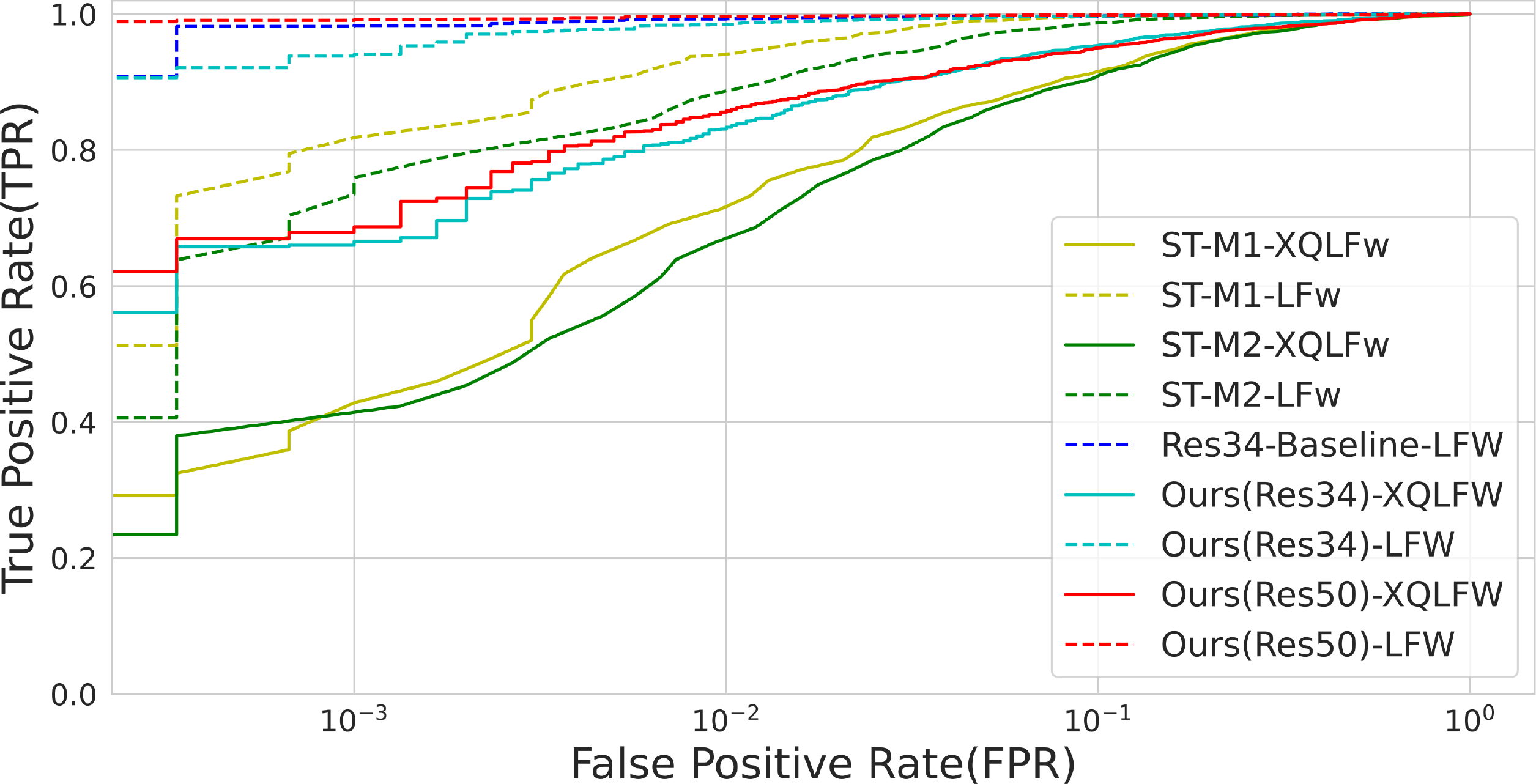}

  \caption{ROC curve of the previous model and our methods on LFW~\cite{LFWTech} and XQLFW~\cite{knoche2021crossquality}.}
  \label{fig:xq_lfw_roc}
\end{figure}

Figure~\ref{fig:xq_lfw_roc} illustrates the ROC curve of the previous model and our methods on LFW and XQLFW.
Similar conclusions could be drawn from Figure~\ref{fig:xq_lfw_roc} as from Table~\ref{tab:xqlfw}.
Meanwhile, our method could achieve a much higher True Positive Rate at low False Positive Rates compared to the previous method, e.g.ST-M~\cite{knoche2021image}, which is a very significant improvement and also very meaningful.
It also might be seen that the results of the ResNet50-based method could obtain a higher performance than the ResNet34-based method~(0.37\% on LFW and 2.15\% on XQLFW).
Our approach demonstrates consistency for the enhancement of backbone and dataset size, which is very meaningful and practical.

\subsection{Ablation Study}

The findings of our ablation study on SCface, XQLFW, and LFW may be found in Table~\ref{tab:ablation}. 
The performance improvement of our proposed \emph{MAug} and $L_{LogExp}$ is consistent and the improvement of each module is very significant

\subsubsection{The effectiveness of $L_{LogExp}$.}
It could be observed that the introduction of a distance function~($L_1$, $L_2$) to constrain the distance between a pair of face embeddings would help the model to improve its performance on LRFR~(about 2\% on XQLFW). 
However, some further performance deterioration on LFW is also observed. Moreover, after replacing $L_1$ and $L_2$ with $L_{LogExp}$, our approach could improve performance on LRFR datasets while reducing performance decay on HRFR. For the case of $p=1$, the increase is 2.55\% on XQLFW, while for the case of $p=2$, it is 2\% (Line 5,6; 8,9). 
With the introduction of Multi-Resolution augmentation, $L_{LogExp}$ can also bring performance gains~(Line 3,7,10) (3\% on SCface and 1\% on XQLFW, while almost unaffected on LFW).
This indicates that introducing direct constraints on features could help the model learn uniform knowledge between LR and HR, but inappropriate constraints may lead to performance degradation on HR, however, our novel distance function could overcome this dilemma. 

\begin{figure}[t]
  \centering
  \includegraphics[width=1.0\linewidth]{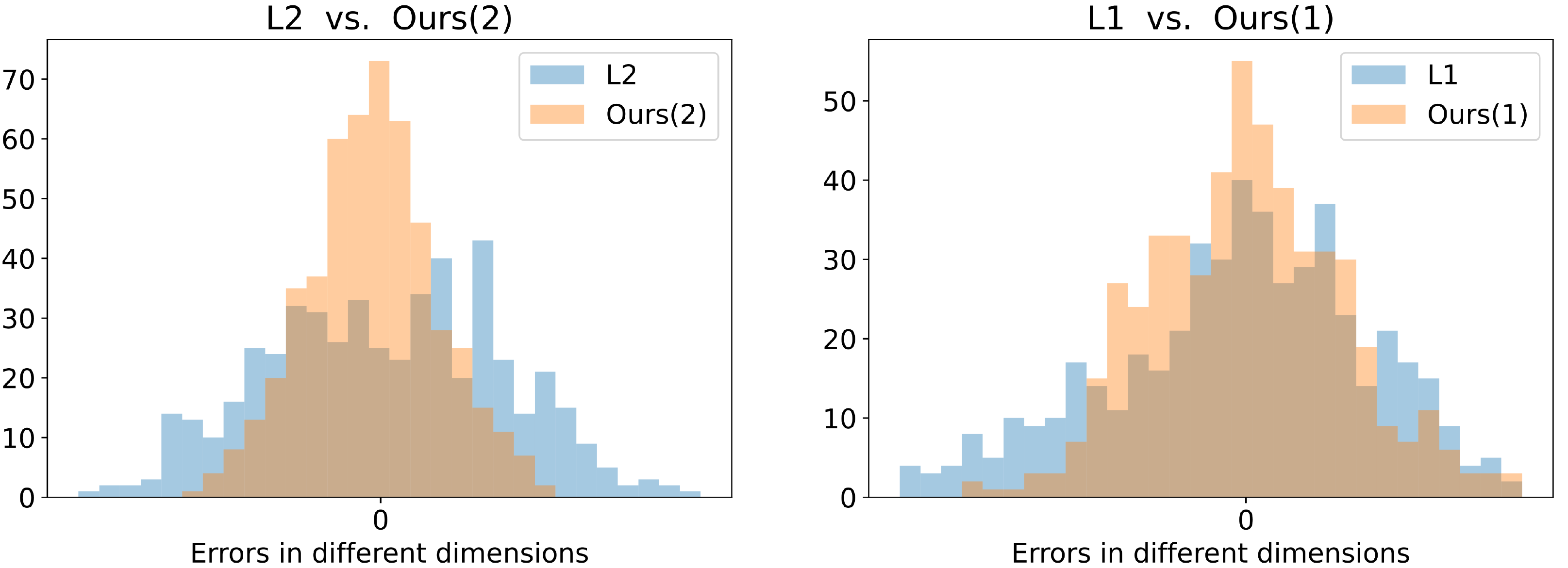}
  \caption{Distribution of the mean values of the errors in different dimensions of the HR and LR features.
  All four models are based on our \emph{MAug} method.}
  \label{fig:feat_err}
\end{figure}

We also selected a set of HR faces, generated the corresponding LR faces, and calculated the distribution of the mean values of the different dimensional errors after extracting features based on four different models.
All four models are based on our \emph{MAug} method while using $L_1$, $L_2$, $L_{LogExp}$~($p=1$) and $L_{LogExp}$~($p=2$), respectively.
As shown in Figure~\ref{fig:feat_err}, the error of our $L_{LogExp}$ is much smaller and can be considered significantly better than the results of $L_1$ and $L_2$ losses.

\begin{table}[t]
  \centering
  \resizebox{\linewidth}{!}{
  \begin{tabular}{l|llll|c|c}
  \toprule
  \multirow{2}{*}{Methods} & \multicolumn{4}{c|}{SCface w/o FT} & \multirow{2}{*}{XQLFW} & \multirow{2}{*}{LFW} \\ \cline{2-5}
                          & $d1$   & $d2$   & $d3$   & Avg  &                        &                      \\ \hline
  Baseline                & 40.3  & 92.8  & 98.5 & 77.2 & 68.95                  & \textbf{99.42}                \\
  Aug                     & 71.5  & 92.8  & 96.3 & 86.8 & 84.33                  & 98.65                \\
  MAug                   & 67.0  & 95.3  & 98.0 & 86.8 & 90.89                  & 98.98                \\
  PAug+Ours(1)           & 60.3  & 95.3  & \textbf{98.8} & 84.8 & 90.35                  & 99.02                \\
  \hline
  Aug+$L_1$ & 67.3  & 94.5  & 94.8  & 85.5  & 85.23  & 98.57  \\
  Aug+Ours(1) & 70.0  & 94.8  & 96.5  & 87.1  & 87.78  & 98.60  \\
  MAug+Ours(1) & 73.0  & \textbf{96.0}  & \textbf{98.8}  & \textbf{89.3}  & \textbf{92.18}  & 99.03  \\
  \hline
  Aug+$L_2$ & 71.8  & 90.0  & 92.8  & 84.8  & 84.87  & 98.50  \\
  Aug+Ours(2) & 72.5  & 92.3  & 94.3  & 86.3  & 86.88  & 98.63  \\
  MAug+Ours(2) & \textbf{74.0}  & 95.5  & 97.8  & \emph{89.1}  & \emph{91.98}  & \emph{99.08}  \\
  \bottomrule
  \end{tabular}
  }
  \caption{Ablation Study based on ResNet34 and CASIA-WebFace. \emph{Baseline} is the HR model with CosFace. \emph{Aug} indicates that we only introduced $14~px$ BICUBIC-based LR faces during training. 
  \emph{MAug} indicates that Multi-resolution augmentation is used in training. \emph{PAug} indicates that the augmentation used in ST-M ($7px$,$14px$,$28px$,$56px$) is used for comparison.
  $L_1$ and $L_2$ represent that we introduced $L_1$  and $L_2$  losses to constrain the embedding of a pair of LR and HR faces, respectively. Ours($x$) indicates that we use $L_{LogExp}$ with $p=x$ as a replacement for $L_1$ or $L_2$.}
\label{tab:ablation}
\end{table}

\subsubsection{The effectiveness of \emph{MAug}.}
Our new Multi-Resolution augmentation strategy could also bring performance improvements. With and without introducing constraint functions, ours \emph{MAug} show performance improvements over \emph{Aug} in both LRFR as well as HRFR~(Line 2,3; 6,7; 9,10).
Significant performance improvements can also be found when comparing with the augmentation~(\emph{PAug}) in ST-M~\cite{knoche2021image}~(Line 4,7) (1.5\% on XQLFW and 4.4\% on SCface).
These indicates that Multi-Resolution augmentation could help the model learn more general knowledge in both LR and HR, helping it to handle the differences between resolution domains and then project a wide range of resolution images to a unified feature space.
This is consistent with our findings in Section~\ref{multiresolu} and therefore supports our conclusion.

\begin{figure}[t]
    \centering
    \includegraphics[width=1.0\linewidth]{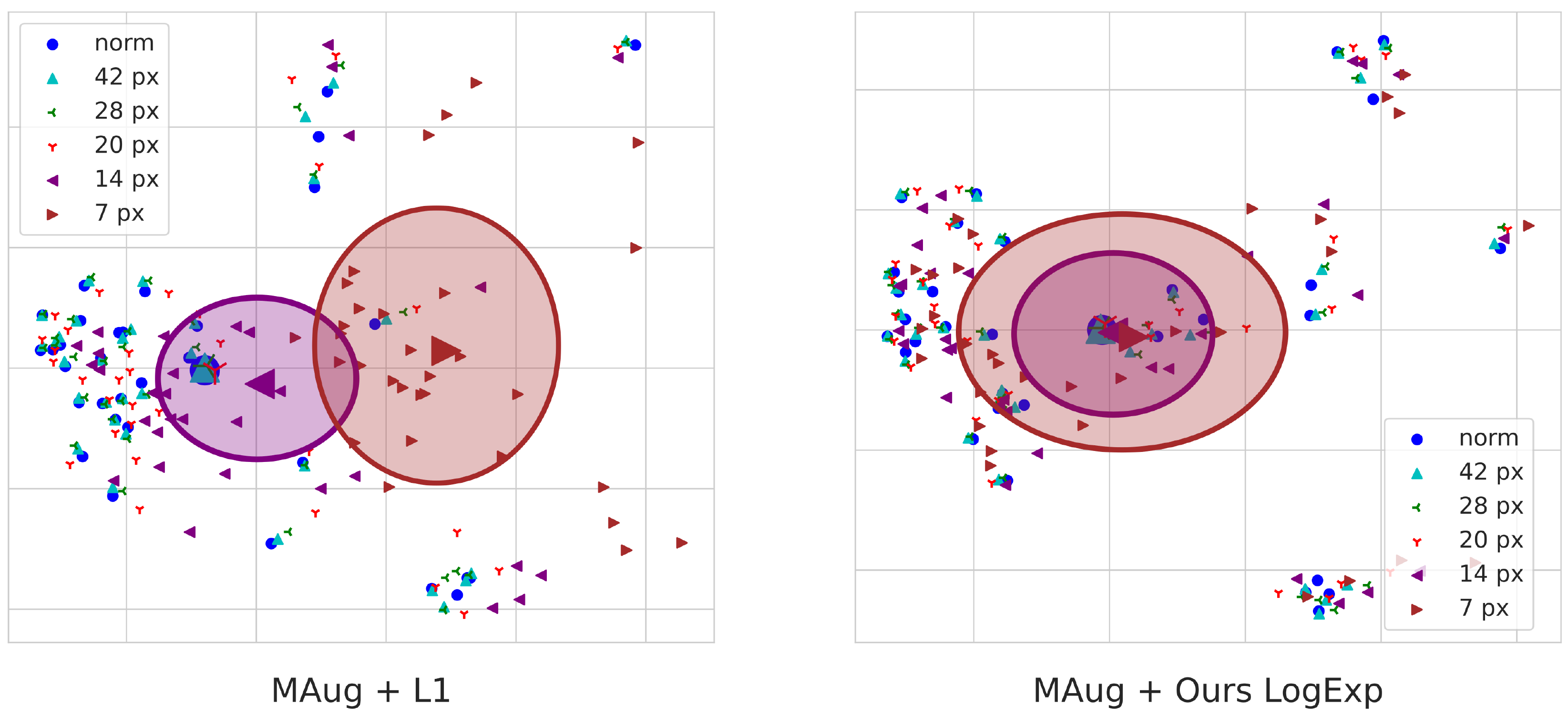}
    \caption{Feature PCA results for models with different contrast loss for the same set of faces at different resolutions~(HR,$42px$,$28px$,$20px$,$14px$,$7px$,). Left: $L_1$, Right: Ours $L_{LogExp}$.
    These two models both utilized our \emph{MAug} method. The bigger point represents the center.}
    \label{fig:pca}
\end{figure}

\subsubsection{Visualization Analysis.}

To show the effectiveness and superiority of our $L_{LogExp}$ more intuitively, we performed a visualization analysis, and the results are shown in Figure~\ref{fig:pca}. 
We selected a set of HR faces and generated corresponding LR faces with 5 different resolutions, after which we observed the distribution of features by PCA~\cite{pearson1901liii}. 
Our proposed $L_{LogExp}$ can indeed optimize the model better and fuse the HR and LR domains better, resulting in a better representation.

\subsubsection{The impact of $\lambda$.}
We also performed more detailed ablation study on $\lambda$.
As shown in Table~\ref{tab:lambda}, the impact of $\lambda$ on performance is not significant. Larger $\lambda$ means stricter constraints, which will bring performance improvement on XQLFW but performance degradation on LFW.
However, excessively strict constraints will harm both. The optimal choice on LR is $\lambda=5$, however in the main text we choose $\lambda=1$ to minimize HR performance degradation, which could also achieve considerable LR performance.

\begin{table}[t]
  \centering
  \resizebox{\linewidth}{!}{
  \begin{tabular}{l|rrrrrr}
  \toprule
  $\lambda$ &0.1& 1     & 3     & 5     & 7     & 10      \\
  \hline
  XQLFW     &91.97 & 92.18 & 92.25 & \textbf{92.45} & 92.31 & 92.17   \\
  LFW       &98.86 & \textbf{99.03} & 98.91 & 98.89 & 98.91 & 98.78   \\
  \bottomrule
  \end{tabular}%
  }
  \caption{The Impact of $\lambda$ on LFW and XQLFW.}
  \label{tab:lambda}
\end{table}%

\subsubsection{Analysis on \emph{MAug} Ratio.}
We also investigate the ratio of different resolutions in multi resolution augmentation as shown in Table~\ref{tab:ablation2}.
The results in Lines 3-5 illustrate that the different resolutions we introduced~($7 px$;$14 px$;$20 px$) are all effective and indispensable. 
This also proves that the difficulty division for resolutions proposed by us is reasonable. 
The lack of \emph{extremely hard}~($7 px$) or \emph{hard}~($14 px$) samples will result in unsatisfactory LR performance (SCface $d1$; XQLFW), while the lack of \emph{semi-hard} samples~($20 px$) will cause a fragmentation of the domain between LR and HR and affect the performance under HR~(SCface $d3$, LFW).
Only by combining resolutions of different difficulty, multi-resolution augmentation can exert the maximum performance.

\begin{table}[t]
  \centering
  \resizebox{\linewidth}{!}{
  \begin{tabular}{c|llll|c|c}
  \toprule
  \multirow{2}{*}{Ratio} & \multicolumn{4}{c|}{SCface w/o FT} & \multirow{2}{*}{XQLFW} & \multirow{2}{*}{LFW} \\ \cline{2-5}
                          & $d1$    & $d2$    & $d3$   & Avg  &                        &                      \\ \hline
  Baseline                & 40.3  & 92.8  & 98.5 & 77.2 & 68.95                  & \textbf{99.42}                \\
  MAug only               & 67.0  & 95.3  & 98.0 & 86.8 & 90.89                  & 98.98                \\
  \hline
  1:1:0                  & \textbf{73.8}   & 94.0   & 96.8   & 88.2   &   \textbf{92.43}                &   98.68              \\
  0:1:1                  & 71.5   & 95.8   & 98.3   & 88.5   &   83.52                &   99.05              \\
  1:0:1                  & 65.3   & 94.8   & 97.8   & 86.0   &   90.53                &   99.02              \\
  \hline
  1:1:1                  & 72.5   & 96.5   & 98.3   & 89.1   &  91.45                 &   98.88              \\
  1:2:1                  & 72.8   & 95.0   & 98.0   & 88.6   &   91.68                &   98.82              \\
  2:1:1                  & 72.8   & \textbf{96.8}   & 97.3   & 88.9   &   92.05                &   98.75              \\
  \hline
  1:1:2                  & 73.0   & 96.0   & \textbf{98.8}   & 89.3   &  92.18                 &  99.03               \\
  \bottomrule
\end{tabular}}
\caption{Ablation Study of the Multi-Resolution Augmentation ratio (7$px$:14$px$:20$px$) based on ResNet34 and $L_{LogExp}$.}

\label{tab:ablation2}
\end{table}

Moreover, similar to the findings of FaceNet~\cite{schroff2015facenet}, we experimentally prove that among these three different resolutions, we need to pay more attention to the \emph{semi-hard} samples~(Lines 7-9).
The introduction of \emph{semi-hard} samples $20~px$ is to better build a bridge between the LR and HR domains, so as to optimize the hard samples more seamlessly. 
However, if pay unnecessarily too much attention to the \emph{hard} and \emph{extremely hard} samples, it will lead to poor performance due to optimization difficulties. Therefore, we finally adopt the ratio of 1:1:2 and verify its performance.

\section{Discussion and Conclusion}

In this paper, we dive into the resolution augmentations and the metrics in LRFR.
We first analyze the impact of different resolution augmentations. Combining the analysis of model gradients, we suggest dividing the low resolution into three difficulties and propose a new Multi-Resolution Augmentation.
We also analyze the popular $L_1$ and $L_2$ distance functions from a gradient viewpoint and propose a novel distance function $L_{LogExp}$ that combines the benefits of both. 
By combining these two techniques, our model could learn more general knowledge in both LR faces and HR faces and boost the performance simultaneously.

\subsubsection{Potential Social Impacts.}
Our experiments use the MS1M dataset, which has been withdrawn by its creator.
However, our usage of MS1M is necessary for a pair comparison to previous works.
Moreover, the subject of this paper is Low Resolution Face Recognition, which may be used in monitoring facilities to obtain face data or for surveillance purposes, etc., without people's knowledge.

\bibliography{refs}

\begin{thebibliography}{54}
\providecommand{\natexlab}[1]{#1}

\bibitem[{Ataer-Cansizoglu et~al.(2019)Ataer-Cansizoglu, Jones, Zhang, and
  Sullivan}]{ataer2019verification}
Ataer-Cansizoglu, E.; Jones, M.; Zhang, Z.; and Sullivan, A. 2019.
\newblock Verification of very low-resolution faces using an
  identity-preserving deep face super-resolution network.
\newblock \emph{arXiv preprint arXiv:1903.10974}.

\bibitem[{Cao et~al.(2018)Cao, Shen, Xie, Parkhi, and
  Zisserman}]{cao2018vggface2}
Cao, Q.; Shen, L.; Xie, W.; Parkhi, O.~M.; and Zisserman, A. 2018.
\newblock Vggface2: A dataset for recognising faces across pose and age.
\newblock In \emph{2018 13th IEEE international conference on automatic face \&
  gesture recognition (FG 2018)}, 67--74. IEEE.

\bibitem[{Chen et~al.(2020)Chen, Kornblith, Norouzi, and
  Hinton}]{chen2020simple}
Chen, T.; Kornblith, S.; Norouzi, M.; and Hinton, G. 2020.
\newblock A simple framework for contrastive learning of visual
  representations.
\newblock In \emph{International conference on machine learning}, 1597--1607.
  PMLR.

\bibitem[{Chen et~al.(2018)Chen, Tai, Liu, Shen, and Yang}]{chen2018fsrnet}
Chen, Y.; Tai, Y.; Liu, X.; Shen, C.; and Yang, J. 2018.
\newblock Fsrnet: End-to-end learning face super-resolution with facial priors.
\newblock In \emph{Proceedings of the IEEE Conference on Computer Vision and
  Pattern Recognition}, 2492--2501.

\bibitem[{Cheng, Zhu, and Gong(2018)}]{cheng2018low}
Cheng, Z.; Zhu, X.; and Gong, S. 2018.
\newblock Low-resolution face recognition.
\newblock In \emph{Asian Conference on Computer Vision}, 605--621. Springer.

\bibitem[{Deng et~al.(2019)Deng, Guo, Xue, and Zafeiriou}]{deng2019arcface}
Deng, J.; Guo, J.; Xue, N.; and Zafeiriou, S. 2019.
\newblock Arcface: Additive angular margin loss for deep face recognition.
\newblock In \emph{Proceedings of the IEEE/CVF Conference on Computer Vision
  and Pattern Recognition}, 4690--4699.

\bibitem[{Fang et~al.(2020)Fang, Deng, Zhong, and Hu}]{fang2020generate}
Fang, H.; Deng, W.; Zhong, Y.; and Hu, J. 2020.
\newblock Generate to adapt: Resolution adaption network for surveillance face
  recognition.
\newblock In \emph{European Conference on Computer Vision}, 741--758. Springer.

\bibitem[{Ge et~al.(2020{\natexlab{a}})Ge, Zhang, Liu, Hua, Zhao, Jin, and
  Wen}]{ge2020look}
Ge, S.; Zhang, K.; Liu, H.; Hua, Y.; Zhao, S.; Jin, X.; and Wen, H.
  2020{\natexlab{a}}.
\newblock Look one and more: Distilling hybrid order relational knowledge for
  cross-resolution image recognition.
\newblock In \emph{Proceedings of the AAAI Conference on Artificial
  Intelligence}, volume~34, 10845--10852.

\bibitem[{Ge et~al.(2018)Ge, Zhao, Li, and Li}]{ge2018low}
Ge, S.; Zhao, S.; Li, C.; and Li, J. 2018.
\newblock Low-resolution face recognition in the wild via selective knowledge
  distillation.
\newblock \emph{IEEE Transactions on Image Processing}, 28(4): 2051--2062.

\bibitem[{Ge et~al.(2020{\natexlab{b}})Ge, Zhao, Li, Zhang, and
  Li}]{ge2020efficient}
Ge, S.; Zhao, S.; Li, C.; Zhang, Y.; and Li, J. 2020{\natexlab{b}}.
\newblock Efficient low-resolution face recognition via bridge distillation.
\newblock \emph{IEEE Transactions on Image Processing}, 29: 6898--6908.

\bibitem[{Ghosh, Singh, and Vatsa(2019)}]{ghosh2019learning}
Ghosh, S.; Singh, R.; and Vatsa, M. 2019.
\newblock On learning density aware embeddings.
\newblock In \emph{Proceedings of the IEEE/CVF Conference on Computer Vision
  and Pattern Recognition}, 4884--4892.

\bibitem[{Grgic, Delac, and Grgic(2011)}]{grgic2011scface}
Grgic, M.; Delac, K.; and Grgic, S. 2011.
\newblock SCface--surveillance cameras face database.
\newblock \emph{Multimedia tools and applications}, 51(3): 863--879.

\bibitem[{Guo et~al.(2016)Guo, Zhang, Hu, He, and Gao}]{ms1m}
Guo, Y.; Zhang, L.; Hu, Y.; He, X.; and Gao, J. 2016.
\newblock Ms-celeb-1m: A dataset and benchmark for large-scale face
  recognition.
\newblock In \emph{European conference on computer vision}, 87--102. Springer.

\bibitem[{He et~al.(2020)He, Fan, Wu, Xie, and Girshick}]{he2020momentum}
He, K.; Fan, H.; Wu, Y.; Xie, S.; and Girshick, R. 2020.
\newblock Momentum contrast for unsupervised visual representation learning.
\newblock In \emph{Proceedings of the IEEE/CVF Conference on Computer Vision
  and Pattern Recognition}, 9729--9738.

\bibitem[{He et~al.(2016)He, Zhang, Ren, and Sun}]{he2016deep}
He, K.; Zhang, X.; Ren, S.; and Sun, J. 2016.
\newblock Deep residual learning for image recognition.
\newblock In \emph{Proceedings of the IEEE conference on computer vision and
  pattern recognition}, 770--778.

\bibitem[{Hong and Ryu(2019)}]{hong2019unsupervised}
Hong, S.; and Ryu, J. 2019.
\newblock Unsupervised face domain transfer for low-resolution face
  recognition.
\newblock \emph{IEEE Signal Processing Letters}, 27: 156--160.

\bibitem[{Huang et~al.(2007)Huang, Ramesh, Berg, and Learned-Miller}]{LFWTech}
Huang, G.~B.; Ramesh, M.; Berg, T.; and Learned-Miller, E. 2007.
\newblock Labeled Faces in the Wild: A Database for Studying Face Recognition
  in Unconstrained Environments.
\newblock Technical Report 07-49, University of Massachusetts, Amherst.

\bibitem[{Huang et~al.(2020)Huang, Shen, Tai, Li, Liu, Li, Huang, and
  Ji}]{huang2020improving}
Huang, Y.; Shen, P.; Tai, Y.; Li, S.; Liu, X.; Li, J.; Huang, F.; and Ji, R.
  2020.
\newblock Improving face recognition from hard samples via distribution
  distillation loss.
\newblock In \emph{European Conference on Computer Vision}, 138--154. Springer.

\bibitem[{Khalid et~al.(2022)Khalid, Awais, Feng, Chan, Farooq, Akbari, and
  Kittler}]{khalid2022npt}
Khalid, S.~S.; Awais, M.; Feng, Z.; Chan, C.~H.; Farooq, A.; Akbari, A.; and
  Kittler, J. 2022.
\newblock NPT-Loss: Demystifying face recognition losses with Nearest Proxies
  Triplet.
\newblock \emph{IEEE transactions on pattern analysis and machine
  intelligence}.

\bibitem[{Knoche, H{\"o}rmann, and Rigoll(2021)}]{knoche2021image}
Knoche, M.; H{\"o}rmann, S.; and Rigoll, G. 2021.
\newblock Image Resolution Susceptibility of Face Recognition Models.
\newblock \emph{arXiv preprint arXiv:2107.03769}.

\bibitem[{Knoche, Hörmann, and Rigoll(2021)}]{knoche2021crossquality}
Knoche, M.; Hörmann, S.; and Rigoll, G. 2021.
\newblock Cross-Quality LFW: A Database for Analyzing Cross-Resolution Image
  Face Recognition in Unconstrained Environments.
\newblock arXiv:2108.10290.

\bibitem[{Li, Liu, and Wang(2019)}]{li2019gradient}
Li, B.; Liu, Y.; and Wang, X. 2019.
\newblock Gradient harmonized single-stage detector.
\newblock In \emph{Proceedings of the AAAI conference on artificial
  intelligence}, volume~33, 8577--8584.

\bibitem[{Li et~al.(2018)Li, Prieto, Mery, and Flynn}]{li2018face}
Li, P.; Prieto, L.; Mery, D.; and Flynn, P. 2018.
\newblock Face recognition in low quality images: a survey.
\newblock \emph{arXiv preprint arXiv:1805.11519}.

\bibitem[{Li et~al.(2019)Li, Prieto, Mery, and Flynn}]{li2019low}
Li, P.; Prieto, L.; Mery, D.; and Flynn, P.~J. 2019.
\newblock On low-resolution face recognition in the wild: Comparisons and new
  techniques.
\newblock \emph{IEEE Transactions on Information Forensics and Security},
  14(8): 2000--2012.

\bibitem[{Li, Tu, and Xu(2021)}]{li2021deep}
Li, P.; Tu, S.; and Xu, L. 2021.
\newblock Deep Rival Penalized Competitive Learning for Low-resolution Face
  Recognition.
\newblock \emph{arXiv preprint arXiv:2108.01286}.

\bibitem[{Li, Tu, and Xu(2022)}]{li2022deep}
Li, P.; Tu, S.; and Xu, L. 2022.
\newblock Deep Rival Penalized Competitive Learning for low-resolution face
  recognition.
\newblock \emph{Neural Networks}.

\bibitem[{Liu et~al.(2017)Liu, Wen, Yu, Li, Raj, and Song}]{liu2017sphereface}
Liu, W.; Wen, Y.; Yu, Z.; Li, M.; Raj, B.; and Song, L. 2017.
\newblock Sphereface: Deep hypersphere embedding for face recognition.
\newblock In \emph{Proceedings of the IEEE conference on computer vision and
  pattern recognition}, 212--220.

\bibitem[{Liu et~al.(2016)Liu, Wen, Yu, and Yang}]{liu2016large}
Liu, W.; Wen, Y.; Yu, Z.; and Yang, M. 2016.
\newblock Large-margin softmax loss for convolutional neural networks.
\newblock In \emph{ICML}, volume~2, 7.

\bibitem[{Lu, Jiang, and Kot(2018)}]{lu2018deep}
Lu, Z.; Jiang, X.; and Kot, A. 2018.
\newblock Deep coupled resnet for low-resolution face recognition.
\newblock \emph{IEEE Signal Processing Letters}, 25(4): 526--530.

\bibitem[{Massoli, Amato, and Falchi(2020)}]{massoli2020cross}
Massoli, F.~V.; Amato, G.; and Falchi, F. 2020.
\newblock Cross-resolution learning for face recognition.
\newblock \emph{Image and Vision Computing}, 99: 103927.

\bibitem[{Meng et~al.(2021)Meng, Zhao, Huang, and Zhou}]{meng2021magface}
Meng, Q.; Zhao, S.; Huang, Z.; and Zhou, F. 2021.
\newblock Magface: A universal representation for face recognition and quality
  assessment.
\newblock In \emph{Proceedings of the IEEE/CVF Conference on Computer Vision
  and Pattern Recognition}, 14225--14234.

\bibitem[{Mudunuri, Sanyal, and Biswas(2018)}]{mudunuri2018genlr}
Mudunuri, S.~P.; Sanyal, S.; and Biswas, S. 2018.
\newblock GenLR-Net: Deep framework for very low resolution face and object
  recognition with generalization to unseen categories.
\newblock In \emph{2018 IEEE/CVF Conference on Computer Vision and Pattern
  Recognition Workshops (CVPRW)}, 602--60209. IEEE.

\bibitem[{Oord, Li, and Vinyals(2018)}]{oord2018representation}
Oord, A. v.~d.; Li, Y.; and Vinyals, O. 2018.
\newblock Representation learning with contrastive predictive coding.
\newblock \emph{arXiv preprint arXiv:1807.03748}.

\bibitem[{Paszke et~al.(2019)Paszke, Gross, Massa, Lerer, Bradbury, Chanan,
  Killeen, Lin, Gimelshein, Antiga, Desmaison, Kopf, Yang, DeVito, Raison,
  Tejani, Chilamkurthy, Steiner, Fang, Bai, and Chintala}]{NEURIPS2019_9015}
Paszke, A.; Gross, S.; Massa, F.; Lerer, A.; Bradbury, J.; Chanan, G.; Killeen,
  T.; Lin, Z.; Gimelshein, N.; Antiga, L.; Desmaison, A.; Kopf, A.; Yang, E.;
  DeVito, Z.; Raison, M.; Tejani, A.; Chilamkurthy, S.; Steiner, B.; Fang, L.;
  Bai, J.; and Chintala, S. 2019.
\newblock PyTorch: An Imperative Style, High-Performance Deep Learning Library.
\newblock In Wallach, H.; Larochelle, H.; Beygelzimer, A.; d\textquotesingle
  Alch\'{e}-Buc, F.; Fox, E.; and Garnett, R., eds., \emph{Advances in Neural
  Information Processing Systems 32}, 8024--8035. Curran Associates, Inc.

\bibitem[{Pearson(1901)}]{pearson1901liii}
Pearson, K. 1901.
\newblock LIII. On lines and planes of closest fit to systems of points in
  space.
\newblock \emph{The London, Edinburgh, and Dublin philosophical magazine and
  journal of science}, 2(11): 559--572.

\bibitem[{Schroff, Kalenichenko, and Philbin(2015)}]{schroff2015facenet}
Schroff, F.; Kalenichenko, D.; and Philbin, J. 2015.
\newblock Facenet: A unified embedding for face recognition and clustering.
\newblock In \emph{Proceedings of the IEEE conference on computer vision and
  pattern recognition}, 815--823.

\bibitem[{Shi et~al.(2020)Shi, Yu, Sohn, Chandraker, and Jain}]{shi2020towards}
Shi, Y.; Yu, X.; Sohn, K.; Chandraker, M.; and Jain, A.~K. 2020.
\newblock Towards universal representation learning for deep face recognition.
\newblock In \emph{Proceedings of the IEEE/CVF Conference on Computer Vision
  and Pattern Recognition}, 6817--6826.

\bibitem[{Singh et~al.(2021)Singh, Nagpal, Singh, and
  Vatsa}]{singh2021derivenet}
Singh, M.; Nagpal, S.; Singh, R.; and Vatsa, M. 2021.
\newblock DeriveNet for (Very) Low Resolution Image Classification.
\newblock \emph{IEEE Transactions on Pattern Analysis and Machine
  Intelligence}.

\bibitem[{Sun(2015)}]{sun2015deep}
Sun, Y. 2015.
\newblock \emph{Deep learning face representation by joint
  identification-verification}.
\newblock The Chinese University of Hong Kong (Hong Kong).

\bibitem[{Terh{\"o}rst et~al.(2021)Terh{\"o}rst, Ihlefeld, Huber, Damer,
  Kirchbuchner, Raja, and Kuijper}]{terhorst2021qmagface}
Terh{\"o}rst, P.; Ihlefeld, M.; Huber, M.; Damer, N.; Kirchbuchner, F.; Raja,
  K.; and Kuijper, A. 2021.
\newblock QMagFace: Simple and Accurate Quality-Aware Face Recognition.
\newblock \emph{arXiv preprint arXiv:2111.13475}.

\bibitem[{Van~der Maaten and Hinton(2008)}]{van2008visualizing}
Van~der Maaten, L.; and Hinton, G. 2008.
\newblock Visualizing data using t-SNE.
\newblock \emph{Journal of machine learning research}, 9(11).

\bibitem[{Wang et~al.(2018)Wang, Cheng, Liu, and Liu}]{wang2018additive}
Wang, F.; Cheng, J.; Liu, W.; and Liu, H. 2018.
\newblock Additive margin softmax for face verification.
\newblock \emph{IEEE Signal Processing Letters}, 25(7): 926--930.

\bibitem[{Wang and Deng(2018)}]{wang2018survey}
Wang, M.; and Deng, W. 2018.
\newblock Deep face recognition: A survey.
\newblock \emph{arXiv preprint arXiv:1804.06655}.

\bibitem[{Wang et~al.(2004)Wang, Bovik, Sheikh, and Simoncelli}]{wang2004image}
Wang, Z.; Bovik, A.~C.; Sheikh, H.~R.; and Simoncelli, E.~P. 2004.
\newblock Image quality assessment: from error visibility to structural
  similarity.
\newblock \emph{IEEE transactions on image processing}, 13(4): 600--612.

\bibitem[{Yang et~al.(2017)Yang, Yang, Gao, and Liao}]{yang2017discriminative}
Yang, F.; Yang, W.; Gao, R.; and Liao, Q. 2017.
\newblock Discriminative multidimensional scaling for low-resolution face
  recognition.
\newblock \emph{IEEE Signal Processing Letters}, 25(3): 388--392.

\bibitem[{Yi et~al.(2014)Yi, Lei, Liao, and Li}]{webface}
Yi, D.; Lei, Z.; Liao, S.; and Li, S.~Z. 2014.
\newblock Learning face representation from scratch.
\newblock \emph{arXiv preprint arXiv:1411.7923}.

\bibitem[{Zangeneh, Rahmati, and Mohsenzadeh(2020)}]{zangeneh2020low}
Zangeneh, E.; Rahmati, M.; and Mohsenzadeh, Y. 2020.
\newblock Low resolution face recognition using a two-branch deep convolutional
  neural network architecture.
\newblock \emph{Expert Systems with Applications}, 139: 112854.

\bibitem[{Zha and Chao(2019)}]{zha2019tcn}
Zha, J.; and Chao, H. 2019.
\newblock Tcn: Transferable coupled network for cross-resolution face
  recognition.
\newblock In \emph{ICASSP 2019-2019 IEEE International Conference on Acoustics,
  Speech and Signal Processing (ICASSP)}, 3302--3306. IEEE.

\bibitem[{Zhang et~al.(2018)Zhang, Zhang, Cheng, Hsu, Qiao, Liu, and
  Zhang}]{zhang2018super}
Zhang, K.; Zhang, Z.; Cheng, C.-W.; Hsu, W.~H.; Qiao, Y.; Liu, W.; and Zhang,
  T. 2018.
\newblock Super-identity convolutional neural network for face hallucination.
\newblock In \emph{Proceedings of the European conference on computer vision
  (ECCV)}, 183--198.

\bibitem[{Zhang et~al.(2016)Zhang, Zhang, Li, and Qiao}]{MTCNN}
Zhang, K.; Zhang, Z.; Li, Z.; and Qiao, Y. 2016.
\newblock Joint face detection and alignment using multitask cascaded
  convolutional networks.
\newblock \emph{IEEE Signal Processing Letters}, 23(10): 1499--1503.

\bibitem[{Zhang et~al.(2020)Zhang, Deng, Wang, Hu, Li, Zhao, and
  Wen}]{Zhang_2020_CVPR}
Zhang, Y.; Deng, W.; Wang, M.; Hu, J.; Li, X.; Zhao, D.; and Wen, D. 2020.
\newblock Global-Local GCN: Large-Scale Label Noise Cleansing for Face
  Recognition.
\newblock In \emph{Proceedings of the IEEE/CVF Conference on Computer Vision
  and Pattern Recognition (CVPR)}.

\bibitem[{Zhao et~al.(2019)Zhao, Gao, Li, and Ge}]{zhao2019low}
Zhao, S.; Gao, X.; Li, S.; and Ge, S. 2019.
\newblock Low-resolution face recognition in the wild with mixed-domain
  distillation.
\newblock In \emph{2019 IEEE Fifth International Conference on Multimedia Big
  Data (BigMM)}, 148--154. IEEE.

\bibitem[{Zhong and Deng(2021)}]{zhong2021face}
Zhong, Y.; and Deng, W. 2021.
\newblock Face Transformer for Recognition.
\newblock \emph{arXiv preprint arXiv:2103.14803}.

\bibitem[{Zhu et~al.(2021)Zhu, Huang, Deng, Ye, Huang, Chen, Zhu, Yang, Lu, Du
  et~al.}]{zhu2021webface260m}
Zhu, Z.; Huang, G.; Deng, J.; Ye, Y.; Huang, J.; Chen, X.; Zhu, J.; Yang, T.;
  Lu, J.; Du, D.; et~al. 2021.
\newblock Webface260m: A benchmark unveiling the power of million-scale deep
  face recognition.
\newblock In \emph{Proceedings of the IEEE/CVF Conference on Computer Vision
  and Pattern Recognition}, 10492--10502.

\end{thebibliography}

\end{document}